\documentclass[11pt]{article}
\usepackage[margin=1in]{geometry}
\usepackage{amsmath,amssymb,bm}
\usepackage{graphicx,booktabs,microtype}
\usepackage[colorlinks=true,linkcolor=blue,citecolor=blue]{hyperref}
\newcommand{\cjsd}{D_{\mathrm{CJS}}}
\newcommand{\Ix}{I_x}

\title{A Shared Encoder Is Not a Shared Task:\\
Conditional Comparison for Deep Expert Pools}
\author{Kentaro Oda\\ Center for Management of Information Technologies, Kagoshima University\\ \texttt{odaken@cc.kagoshima-u.ac.jp}}
\date{}

\begin{document}
\maketitle

\begin{abstract}
Sharing representations does not make head compatibility equivalent to
mechanism compatibility: head exchange asks whether a head
\emph{extrapolates}; the conditional comparison developed here asks whether
the \emph{mechanism changed}.
A popular recipe for comparing tasks in deep continual learning trains a
shared encoder with per-task heads and cross-evaluates the heads at
representation level: if the swapped head performs well, the tasks are deemed
compatible. We show this deep exchange score can inherit, unchanged, the central
confound of its shallow ancestor: under pure input shift with an identical
labeling mechanism ($15^\circ$--$90^\circ$ rotations of MNIST with labels
fixed), the frozen encoder maps the shifted inputs to unpopulated regions of
the embedding space, the swapped head fails there, and the score inflates from
$0.02$ to $0.80$---indistinguishable from a genuine mechanism change of
comparable magnitude. Mahalanobis-style representation novelty (a member of the
representation-novelty family used by expansion triggers; we do not claim to
reimplement any full system such as SEMA) has the opposite blind spot: it is flat ($\approx$const) across label-permutation
drifts of any magnitude, because permuting labels moves no embedding.
We transplant the conditional Jensen--Shannon discrepancy into the embedding
space: two small MLP discriminators on $[\phi(x)]$ and $[\phi(x),y]$ deliver a
functional axis that stays in $[-0.01,0.01]$ across all rotations while
tracking label-permutation drift mass, and a covariate axis that absorbs the
input shift---the exact chain-rule decomposition holds for the induced
representation-space distributions (the original-mechanism reading
additionally requires the common-factoring assumption of Sec.~2). Built into a
mixture-of-heads system with spawn/reuse gating, the two-axis rule attains the
best decision quality (false-spawn $0.044$, no missed concepts beyond one
borderline seed) with $2.7$ heads where the deep exchange rule spawns $11.3$;
we also chart where simple loss-jump triggers suffice (large abrupt drifts)
and where they fail (gradual and recurring regimes). As an application, the
same embedding-space functional axis separates semantic novelty from
photometric shift on the CIFAR-10 GCD split at AUROC $0.99$, where MSP,
Energy, and Mahalanobis all sit at chance. All findings replicate on
\emph{two} frozen foundation backbones (supervised ImageNet-21k ViT-B/16
and self-supervised DINOv2) over CIFAR-100: the exchange score inflates to
$0.18$--$0.27$ under rotations while the embedding functional axis stays
within $\pm0.002$ and tracks permutation drift mass, and the chunk-level
functional axis separates novel from photometric chunks at AUROC $0.98$
where per-input scores (including KNN-OOD) sit at $0.65$--$0.68$. In a
real adapter pool with recurrence, an actual SEMA-style novelty trigger
never fires on mechanism changes (single-adapter collapse, $0.686$),
while the two-axis gate reaches $0.847$ with the oracle number of
adapters and $100\%$ reuse of the recurring concept's adapter.
\end{abstract}

\section{Introduction}
Deep continual learning systems increasingly share a frozen or slowly-updated
encoder and manage a growing set of task heads or adapters; the decision to
\emph{reuse} a head, \emph{spawn} a new one, or \emph{merge} two is delegated
to a task-comparison score. Two common score families are: cross-evaluated
heads (the representation-level version of exchange scores such as CLS), and
representation novelty (Mahalanobis or density models on embeddings, as in
SEMA-style expansion triggers). We show both are one-axis instruments, in
opposite directions, and that the failure can be architectural rather than
statistical---sharing an encoder alone need not remove it:
sharing the encoder does not repair it, because a frozen encoder is itself a
covariate-shift amplifier for off-distribution inputs.

\section{The deep confound, measured}
\begin{figure}[t]
\centering
\includegraphics[width=\linewidth]{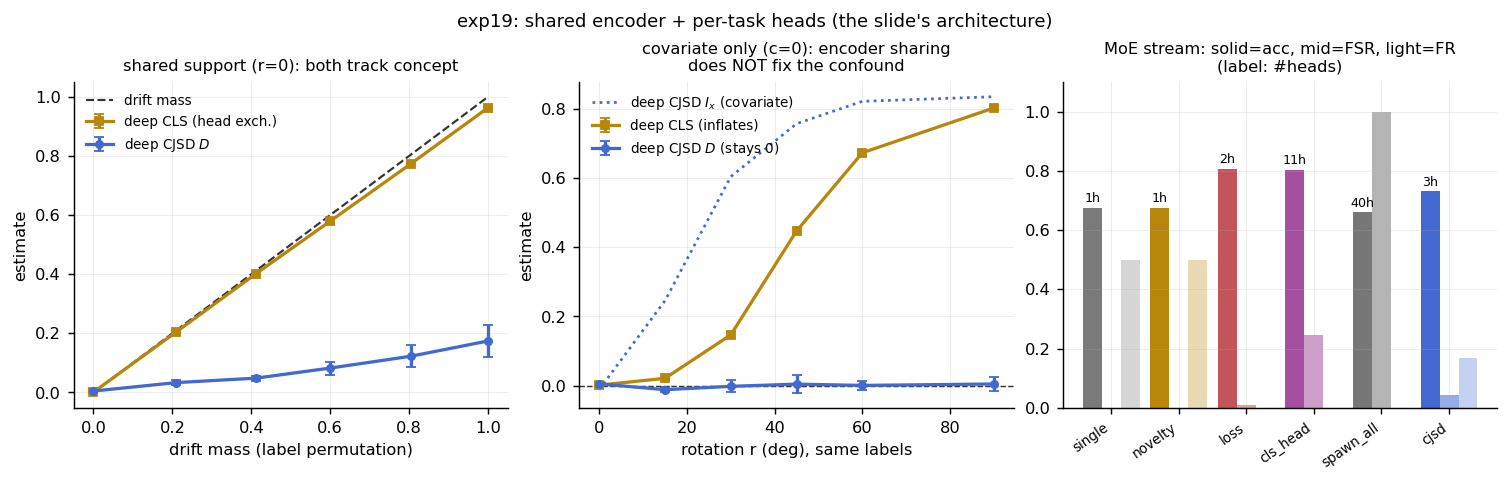}
\caption{Left: under shared support both the deep exchange score and the
embedding CJSD track label-permutation drift mass. Middle: under pure input
rotation (labels fixed) the exchange score inflates to $0.80$ while the
embedding CJSD stays at zero and the covariate axis ($\Ix$, dotted) absorbs
the shift. Right: mixture-of-heads stream---accuracy (solid), false spawns
(mid), missed concepts (light); annotations give final head counts.}
\label{fig:deep}
\end{figure}
Setup: encoder trained on plain MNIST, then frozen; task $B$ applies rotation
$r$ (covariate axis) and a cyclic permutation of $c$ classes (mechanism axis
with drift mass $c/10$); per-task linear/MLP heads on the $128$-d embedding.
Under shared support the exchange score is an excellent drift-mass estimator
(slope ${\approx}1$); under pure rotation it reports $0.15$ at $30^\circ$ and
$0.80$ at $90^\circ$ with the mechanism untouched (Fig.~\ref{fig:deep},
middle). Mahalanobis novelty is the mirror image: monotone in $r$, exactly
flat in $c$. The embedding CJSD, computed by two MLP discriminators on
$[\phi(x)]$ and $[\phi(x), \mathrm{onehot}(y)]$, keeps the functional axis in
$[-0.013,0.012]$ across all rotations while tracking $c$ (attenuated by
discriminator capacity; the shrinkage was downward in all our runs), and its
covariate axis rises $0.25\to0.84$ (all $\Ix$ and $\cjsd$ values in this
paper are normalized by $\ln 2$). A caution is required here:
the chain rule holds exactly for the \emph{induced} distributions on
$(\phi(X),Y)$, but the covariate-null interpretation with respect to the
original mechanism does not transfer for free---for a non-injective $\phi$,
$P_A(Y\mid X)=P_B(Y\mid X)$ does not imply
$P_A(Y\mid\phi(X))=P_B(Y\mid\phi(X)) $ (a two-point counterexample
suffices). Task-wise sufficiency alone ($Y\perp X\mid\phi(X)$ under each
task separately) is \emph{not} enough: it permits different
representation-level conditionals $g_A\ne g_B$, so a common original
mechanism need not remain common after collapse. The condition that does
transfer is a \emph{common factoring}: there exists a single
$g$ with $P_A(Y\mid X{=}x)=P_B(Y\mid X{=}x)=g(Y\mid\phi(x))$ on the union
support---then $P_A(Y\mid\phi(X))=P_B(Y\mid\phi(X))=g$ there, and the
embedding covariate-null follows. Whether a given encoder satisfies common factoring is an \emph{empirical
compatibility assumption}---label predictiveness alone does not establish
it under a shifted task; our experiments are consistent with it holding
approximately for this encoder, and in the rotation experiments the
embedding functional axis is additionally protected by vacuity (the
weighted-JS weight vanishes off-overlap). We state the common-factoring
condition as an assumption, not a theorem-free property.

\section{Two-axis gating for mixtures of heads}
We instantiate a mixture-of-heads system (frozen encoder; heads as experts;
routing by recent loss) and compare six spawn policies on a stream with
ground-truth mapping identities (plain / permuted / plain / rotated-plain /
permuted / plain phases): single head, spawn-always, Mahalanobis novelty,
loss-jump, deep-exchange rule, and the two-axis CJSD rule.
Findings (Fig.~\ref{fig:deep}, right):
representation novelty \emph{cannot see} the permuted concept (labels move no
embedding), collapsing to the single-head behavior (missed-concept rate
$0.5$); the deep-exchange rule over-spawns on the rotated phase (false-spawn
$0.246$, $11.3$ heads); loss-jump excels here because every real switch is
large and abrupt ($0.807$ accuracy)---the regime chart from our streaming
study predicts exactly this, and its converse: on gradual and recurring
regimes (INSECTS-reoccurring) loss triggers lose $5$ accuracy points to
two-axis gating. The CJSD rule holds the best joint decision profile
(false-spawn $0.044$, heads $2.7$) at a $0.07$ accuracy cost from conservative
deferral in this fast-switching stream---the tunable price of evidence-gated
decisions.

\section{Application: semantic novelty vs.\ photometric shift}
\begin{figure}[t]
\centering
\includegraphics[width=0.62\linewidth]{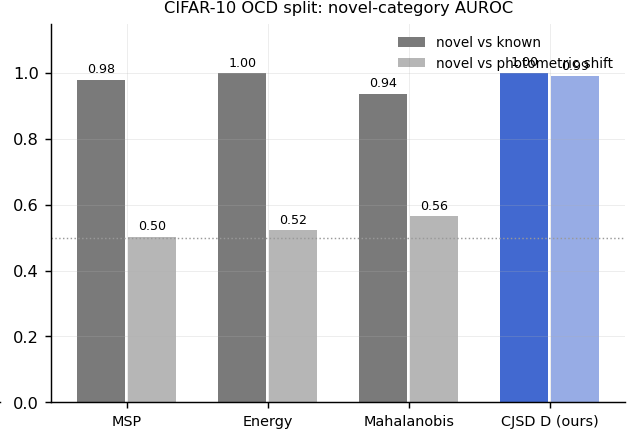}
\caption{CIFAR-10 GCD split. Detecting novel categories against
clean known-class chunks is easy for every OOD score; distinguishing them from
photometric shifts of known classes is at chance for MSP/Energy/Mahalanobis
and near-perfect for the embedding functional axis.}
\label{fig:ocd}
\end{figure}
On the CIFAR-10 GCD split (known classes 0--4), chunks are clean-known,
photometrically-jittered known, or half-novel. Our monitor operates in the
\emph{supervised-batch} regime (chunk labels available, novel labels folded
into an ``other'' bucket), whereas MSP/Energy/Mahalanobis are label-free
single-input scores; the comparison shows what label-free scores structurally
cannot separate, not that they fail at their own task. All standard OOD scores detect
novelty against clean chunks (AUROC $0.94$--$1.00$) and all fail to separate
novelty from photometric shift ($0.50$--$0.57$); the embedding functional
axis achieves $0.99$ while its covariate companion correctly claims the
photometric chunks (Fig.~\ref{fig:ocd}). For deployment this is the
difference between opening a new class and adjusting a normalization layer.
A structural remark on the label requirement: replacing chunk labels with
pseudo-labels that are a \emph{deterministic function of the same inputs}
(e.g.\ the encoder's own predictions $g(x)$) cannot work even in principle,
because such a function carries no information beyond $x$
($I(Z;g(X)\mid X)=0$). The impossibility is scoped to this setting: labels
derived from side information not contained in $x$ (a second modality,
human raters, delayed outcomes) are genuine $y$'s and are not covered by
the argument. Within it, a label-free functional axis is impossible, and
unlabeled novelty detection legitimately reduces to the covariate axis.

\section{Foundation-encoder scale}
The same protocol at modern scale: a frozen ViT-B/16 pretrained on
ImageNet-21k (timm), CIFAR-100 with $50$ known classes, $6000$ images per
side, two MLP discriminators on the $768$-d embeddings. The findings
transfer unchanged. \emph{Confound:} under pure input rotations
($15^\circ$--$90^\circ$, labels intact) the exchange score (cross-evaluated
linear heads) inflates from its null value $0.0005$ to $0.065$--$0.179$,
while the embedding functional axis stays at $|\cjsd|\le 0.001$; under
label permutations of $c$ classes the functional axis tracks drift mass
($0.149/0.371/0.780$ for mass $0.2/0.5/1.0$, the familiar capacity
shrinkage) and the exchange score does too---at this scale, as before, the
exchange score cannot tell the two situations apart and the two-axis
decomposition can. Two honest observations: rotations of $15^\circ$--$60^\circ$
drive the (normalized) covariate axis to $\widehat\Ix\approx0.99$---resampling artifacts
make the embedding supports nearly disjoint, so a lifecycle layer would
declare these pairs \emph{incomparable} rather than reusable (the designed
behavior; $90^\circ$, an artifact-free axis-aligned rotation, sits at
$\Ix=0.48$); and on identical-support pairs the overfit $T_1$ yields a
vacuous negative $\widehat\Ix$ (held-out cross-entropy above $\ln 2$),
which we report as-is and clamp only at the decision layer.
\emph{GCD separation:} with $100$ fine-grained classes the per-input OOD
scores degrade to AUROC $0.64$--$0.70$ and---the structural point---score
novel-class chunks and photometrically-shifted chunks almost identically
(novel-vs-photometric $0.65$--$0.68$); the chunk-level functional axis puts
novel chunks at $\cjsd=0.063\pm0.020$ against $-0.003\pm0.016$ for
photometric and $-0.007\pm0.019$ for clean chunks: novel above photometric
in $10/10$ repetitions, chunk-level AUROC $0.98$.

\emph{Real adapters, a real expansion trigger, and recurrence.} Replacing
linear heads by genuine \emph{adapter} modules (residual bottleneck MLPs,
$768\to64\to768$, per-concept, gradient-trained) and adding a recurrence
phase (the permuted concept returns late in the stream) lets us compare
against a null-calibrated SEMA-\emph{style} representation-novelty
trigger (we implement the trigger family---diagonal-Gaussian NLL under
each adapter's input statistics with a null-calibrated threshold---not the
full released SEMA system): expand when the minimum novelty is exceeded. Over the
8-phase stream (3 seeds) the novelty trigger \emph{never fires}---labels
move no embeddings, so it collapses to a single adapter and misses both
mechanism changes (accuracy $0.686$). The two-axis gate attains $0.847$
with exactly $3$ adapters, zero false expansions, zero missed concepts,
and \emph{$100\%$ reuse of the recurring concept's original adapter};
loss-jump matches on this abrupt stream ($0.848$), and spawn-always pays
$40$ adapters for $0.801$. All policies consume identical training
invocations ($40$), so the training-call budget is matched and the
performance differences arise from the routing/expansion decisions.

\emph{A second backbone.} All pair-level findings replicate on
self-supervised DINOv2 (ViT-B/14, $518\times518$ input resolution):
rotations inflate the
exchange score to $0.068$--$0.272$ (null $-0.001$) while the embedding
functional axis stays within $\pm0.002$; label permutations track drift
mass ($0.080/0.242/0.524$ for mass $0.2/0.5/1.0$); and novel-class chunks
separate from photometric chunks in $6/6$ repetitions. The confound and
its repair are properties of frozen-encoder pipelines, not of one
pretraining recipe.

\emph{Head lifecycle at scale.} The mixture-of-heads stream also transfers:
40 chunks over the phases clean $\to$ 25-class permutation $\to$ clean
$\to$ photometric $\to$ novel classes $\to$ clean (3 seeds), MLP heads on
the frozen ViT embeddings, cross-fitted logistic discriminators with the
head's label-log-likelihood supplied as the sufficient-statistic feature.
The single-look two-axis gate achieves \emph{perfect} decision quality:
zero false spawns (photometric and recurring clean chunks reused), zero
missed concepts (permutation and novel phases each spawn on their first
chunk, where the functional axis jumps to $0.69$ and $0.34$ against a
$\le0.01$ clean baseline), and exactly $3$ heads---one per true
concept---at prequential accuracy $0.849$. The loss-jump trigger matches
this on the present stream (all switches are large and abrupt, the regime
our streaming study predicts it handles), while chunk-level Mahalanobis
novelty on the same embeddings misses \emph{both} mechanism changes
(single-head behavior, accuracy $0.709$): the one-axis blindness is not an
artifact of small encoders.

\section{Practical notes}
In our experiments discriminator under-training shrank the functional axis
toward zero (the sign of the misspecification bias is uncontrolled in
general; see the companion paper); tiny per-chunk discriminators need
${\sim}25$ epochs, and
appending the encoder's softmax to the discriminator features supplies the
label-likelihood statistic that low-capacity discriminators otherwise fail to
learn. Duplicated inputs across the two sides of a comparison create
twin-copy memorization leakage and must be avoided. Under multiple heads,
comparisons are shortlisted by recent loss with no measured decision change.

\section{Related work}
SEMA and MoE-adapter expansion trigger on representation novelty; CLS-style
scores cross-evaluate swapped predictors; OOD scores (MSP, Energy,
Mahalanobis) rank single inputs. All are one-axis devices. Concurrent work on
conditional two-sample testing provides the hypothesis-testing view; our
contribution is the embedding-space decomposition and its use as the decision
layer of a deep expert pool.

\section{Limitations}
The encoder is frozen after warm-up and trained without invariance to the
studied transformations; an augmentation-invariant or foundation-model encoder
may shrink the exchange-score confound, so our title claim is existence
(sharing an encoder does not \emph{by itself} fix it), not universality.
Co-adapting the encoder online would couple the axes and change the induced
conditionals over time (representation-version alignment is required).
Pair-level \emph{and} head-lifecycle results now cover a frozen
ImageNet-21k ViT-B/16 (Sec.~5); the ViT lifecycle stream is shorter
(40 chunks) than the MNIST streams, and modern GCD systems (e.g.\ Vaze et
al.'s semi-supervised pipeline, OpenCon) are not run end-to-end---our
comparison targets the score families their expansion decisions rest on,
not full systems.

\end{document}